\documentclass[USenglish,twocolumn]{article}
\usepackage[utf8]{inputenc}
\usepackage[big,online]{dgruyter}
\usepackage[sort&compress,square,numbers]{natbib}
\usepackage{times}

\title{Predictive Multi-Landmark OCT Tracking for Increased Motion Robustness}
\runningtitle{Predictive Multi-Landmark OCT Tracking}
\begin{document}

  \DOI{10.1515/}
  \openaccess
  \pagenumbering{gobble}



\author*[1,2]{K. Reuter}
\author[1]{S. Guttikonda}
\author[1]{C. U. Karekar} 
\author[2]{C. Betz} 
\author[1]{A. Schlaefer}
\runningauthor{K.Reuter et al.}

\affil[1]{\protect\raggedright Institute of Medical Technology and Intelligent Systems, Hamburg University of Technology, Hamburg, Germany, e-mail: konrad.reuter@tuhh.de}
\affil[2]{\protect\raggedright Department of Otorhinolaryngology, Head and Neck Surgery and Oncology, University Medical Center Hamburg-Eppendorf, Hamburg, Germany}

\abstract{
Optical coherence tomography is a promising modality for markerless motion tracking due to its high spatial resolution and inherent depth perception. However, existing OCT-based tracking approaches are limited in terms of trackable velocity, particularly when multiple landmarks are tracked sequentially for 6D pose estimation. In this work, we present a predictive tracking approach that propagates positional updates between multiple tracked landmarks to obtain a global pose prediction. This enables more robust tracking under high velocities. Our results demonstrate RMSEs below 1~mm for velocities up to 100~mm/s and up to nine consecutively tracked landmarks, highlighting the potential of global motion propagation and prediction for improving the robustness of OCT-based tracking.
}


\maketitle

\section{Introduction}

Precise real-time motion tracking has the potential to support a wide range of biomedical applications, including motion compensation, surgical navigation, and instrument tracking \cite{tracking}. Motion tracking methods are generally divided into marker-based and markerless approaches. Marker-based methods typically achieve high tracking precision but are often impractical for tissue tracking due to the difficulty of attaching markers to biological tissue. Consequently, markerless approaches have gained increasing attention, including stereo RGB and RGB-D camera systems. However, these methods generally fail to match the precision of marker-based systems, particularly under challenging intraoperative conditions involving specular reflections or changing illumination \cite{tracking_review}.

As an alternative to predominantly RGB-based approaches, Schlüter et al. proposed a markerless tracking system based on optical coherence tomography (OCT) \cite{6D_OCT}. OCT is less susceptible to reflections, provides high spatial resolution, and inherently captures depth information, enabling accurate depth estimation and the use of subsurface features. However, its field of view is typically limited to only a few millimeters per axis. To compensate for this limitation, the scanning position is actively adjusted during tracking using galvo mirrors for lateral motion compensation and a movable reference arm for depth adjustment. Using this setup, submillimeter tracking accuracy on biological tissue was demonstrated. By simultaneously tracking three or more sequentially updated landmarks, full 6D pose estimation becomes possible \cite{6D_OCT}.

Despite these promising results, the trackable velocities remain limited. Previously employed template-matching approaches, such as MOSSE \cite{MOSSE}, rely on sufficient overlap between the stored template and the currently acquired volume. Once the relative displacement between consecutive acquisitions becomes too large, the overlap is insufficient for reliable motion estimation.

As motion is typically temporally correlated, the history of tracked landmark positions can be utilized to predict the landmark position in the subsequent time step. By compensating for the expected motion, predictive tracking has the potential to increase the maximum trackable velocities.

The problem becomes more severe when multiple landmarks must be tracked simultaneously. Since the landmarks are acquired sequentially, the effective acquisition rate per landmark decreases as the number of tracked landmarks increases, reducing the maximum trackable velocities.

However, landmarks are typically located in close spatial proximity and therefore exhibit similar motion. Consequently, motion information obtained from one landmark can be used to predict the motion of neighboring landmarks. Prior work used an independent tracker instance for each landmark and therefore implicitly assumed fully independent motion.

Based on these observations, we propose a predictive multi-landmark tracking approach that jointly incorporates information from multiple landmarks to estimate a global pose change. By propagating motion information from each acquired volume across all landmarks and predicting their positions at the next timestep, the proposed method enables tracking at considerably higher velocities and with more landmarks. We evaluate the method on OCT scans of porcine skin acquired under simulated motion profiles and demonstrate improved performance compared to the previous approach.

\begin{figure*}
    \centering
    \includegraphics[width=\textwidth]{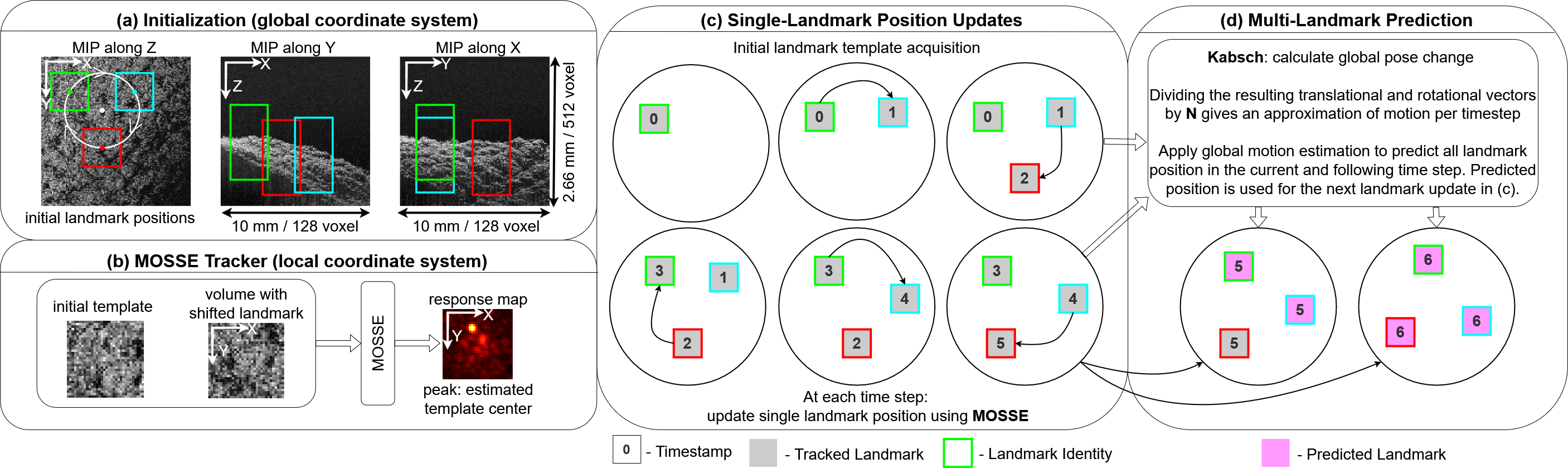}
    \caption{Overview on the proposed method. (a) shows the maximum intensity projections (MIP) for a porcine skin sample with three initial landmarks marked. In (b), the basic concept of MOSSE is shown. The position of the initial template in the current volume is predicted as a peak in the response map. (c) visualizes the update logic for the individual landmarks. Per time step, a single landmarks position is updated using its respective MOSSE instance. (d) depicts the prediction step. For two sets of N landmarks, the global pose change is estimated using the Kabsch algorithm, and subsequently used to predict all landmark positions in the current and following time step.}
    \label{fig:method}
\end{figure*}

\section{Methods}

\subsection{Tracking Method}

Our method employs an individual tracker instance for each tracked landmark. Throughout this work, we use MOSSE, as it has proven to be a fast and accurate baseline for OCT-based tracking. However, the proposed framework is not limited to MOSSE and can readily incorporate other trackers. We assume that all landmarks belong to the same rigid structure. During tracking, landmark positions are updated sequentially using OCT volumes acquired at their currently estimated locations.

Each tracker estimates the position of its target landmark within the acquired OCT volume. Combining this local position estimate with the corresponding acquisition position transforms all landmark positions into a shared global reference frame. In the physical system, acquisition positions are given in motor steps and converted into voxel coordinates using the calibration described in \cite{6D_OCT}.

At each time step, the position of the currently active landmark is updated using its corresponding MOSSE tracker. Subsequently, the transformation between the current and previous landmark configurations is estimated using the Kabsch algorithm \cite{Kabsch}. This transformation resembles the approximate motion over N steps, where N denotes the number of tracked landmarks. By dividing the estimated translational and rotational vectors by N, an approximation for pose change per time step is acquired. Under the assumption of approximately constant velocities in small time intervals, we use this information to predict the positions of all landmarks in the current and the next time step. An overview of the proposed workflow is shown in Figure \ref{fig:method}.

\subsection{Dataset Acquisition}

For data acquisition, the scan head of a Telesto OCT system (Thorlabs, USA) is mounted on an IRB-120 robotic arm (ABB, Switzerland). Following the procedure described in \cite{TCP}, a TCP calibration between the robot end-effector and the OCT field of view is performed.

Nine different regions on porcine skin samples are recorded. For each region, the acquisition angle around the x- and y-axes is varied between -5$^\circ$, 0$^\circ$, and 5$^\circ$, resulting in a total of nine OCT volumes per region. All volumes are acquired at a resolution of 128 $\times$ 128 $\times$ 512 voxels, corresponding to a physical size of 10 $\times$ 10 $\times$ 2.66 mm.

Using the TCP calibration, all recorded volumes are transformed into a common coordinate system defined by the robot base frame. Residual misalignments between transformed volumes are subsequently corrected using iterative closest point (ICP) registration.

For each region, a circle with a radius of 32 voxels is manually defined in the maximum-intensity projection along the depth axis. Landmark positions are then sampled equidistantly along this circle. For each landmark, the tissue surface is detected to determine the initial depth position for subvolume extraction. A sample configuration for three landmarks can be seen in Figure \ref{fig:method} (a).

\subsection{Simulation Environment}

A motion simulation environment is implemented for evaluation. At each time step, an OCT volume is generated as a linear combination of two randomly selected real OCT volumes, simulating natural speckle variation. The volume is virtually moved and subvolumes of size 32 $\times$ 32 $\times$ 256 voxels are extracted for tracking. A volume rate of 832~Hz is assumed, corresponding to an acquisition time of approximately 1.2~ms per volume. The latency of the employed tracker was also taken into account.

The galvo mirror adjustment time is modeled as the sum of a constant small-step response and a linear component for angular changes exceeding the typical small-step range (1$^\circ$). The small-step response time is set to 0.45 ms, while the linear factor is set to 0.05~ms/deg.

Motion of the reference arm is modeled using sinusoidal motion profiles with velocity and acceleration limits corresponding to those of a V-855 high-speed linear stage (PI, Germany), with a maximum velocity of 4~m/s and a maximum acceleration of 50~m/s$^2$.  

\subsection{Evaluation Protocol}

In a first experiment, the required overlap between consecutive OCT volumes for reliable tracking is investigated. For each landmark, two random OCT volumes are selected. From the first volume, a subvolume centered at the landmark position is extracted. In the second volume, the extraction position is shifted along one of the two lateral axes. The shift magnitude is varied between $\pm$1 and $\pm$16 voxels in both lateral directions. For each shift value, a standard MOSSE tracker is used to estimate the induced shift.

\begin{figure}[ht]
    \centering
    \includegraphics[width=1.0\linewidth]{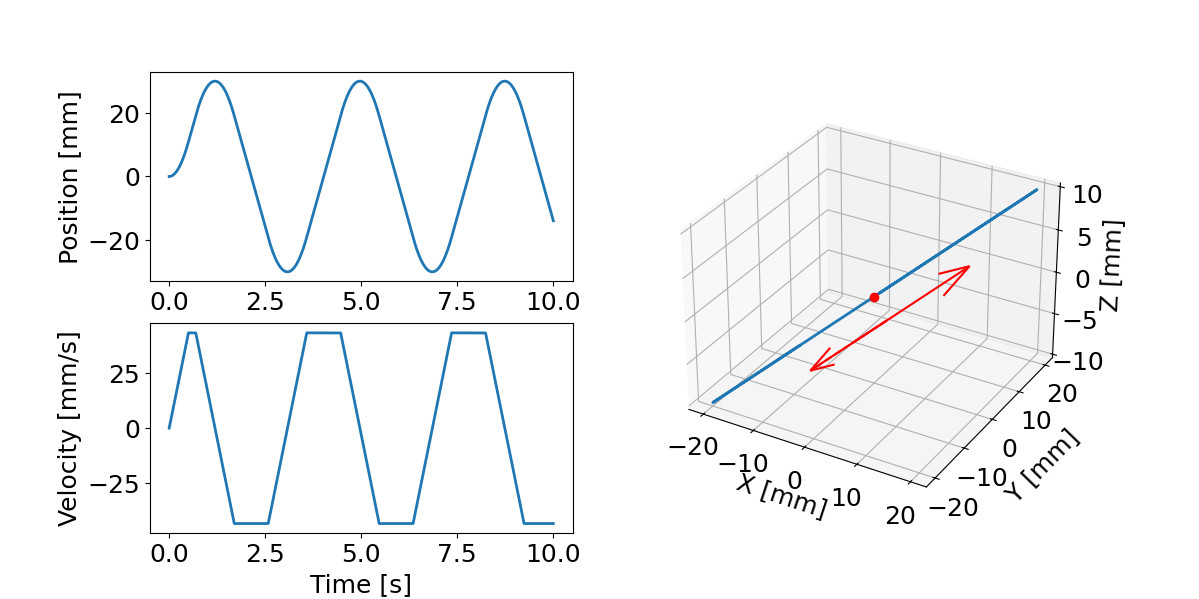}
    \caption{Example trajectory for tracking evaluation at 40 mm/s.}
    \label{fig:trajectory}
\end{figure}

Subsequently, the tracking performance of the baseline approach using an independent tracker per landmark is compared with that of the proposed method. The number of simultaneously tracked landmarks is set to 3, 5, 7, and 9. Motion amplitude is limited to 40~mm for the lateral dimensions (X, Y) and 20~mm for depth dimension (Z), to account for the smaller physical FOV in that direction. Each trajectory has a duration of 30 seconds and continuously alternates between the minimum and maximum values along all axes. Velocity along the diagonal path is varied from 10~mm/s up to 100~mm/s. Acceleration at the turning points is set to 100~mm/s$^2$. The trajectory is visualized in Figure \ref{fig:trajectory}.

Each MOSSE tracker is initialized with a learning rate of 0.02, a Gaussian sigma value of 2.0, and a regularization parameter of $\lambda =$ 10$^{-3}$.

Tracking performance is evaluated using the root mean square error (RMSE). A single position estimate is obtained from the tracked landmark positions using the centroid of all landmarks.

\section{Results}

\begin{figure}[ht]
\centering
\includegraphics[width=1.0\linewidth]{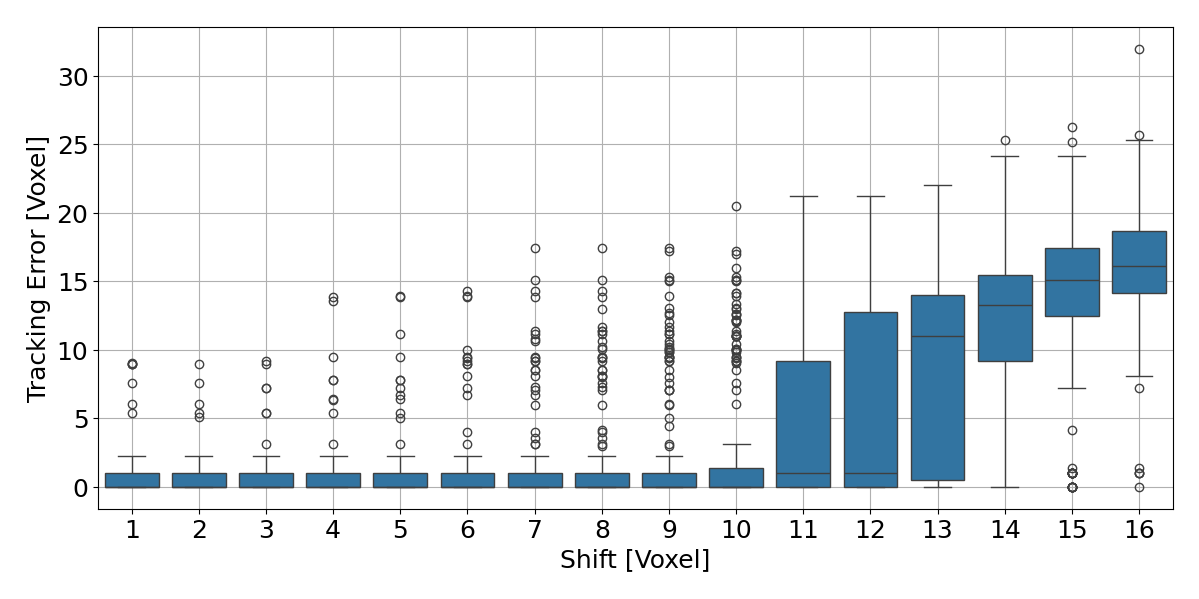}
\caption{Errors in estimating lateral shifts.}
\label{fig:max_shift}
\end{figure}

The results shown in Figure \ref{fig:max_shift} indicate that lateral displacements of up to approximately 11 voxels can be tracked reliably, although the number of outliers already begins to increase before reaching this threshold. Beyond this point, the tracking error increases substantially. From displacements of 13 voxels onward, the median error becomes comparable to the induced displacement, indicating that reliable tracking is no longer feasible.

\begin{figure}[ht]
\centering
\includegraphics[width=1.0\linewidth]{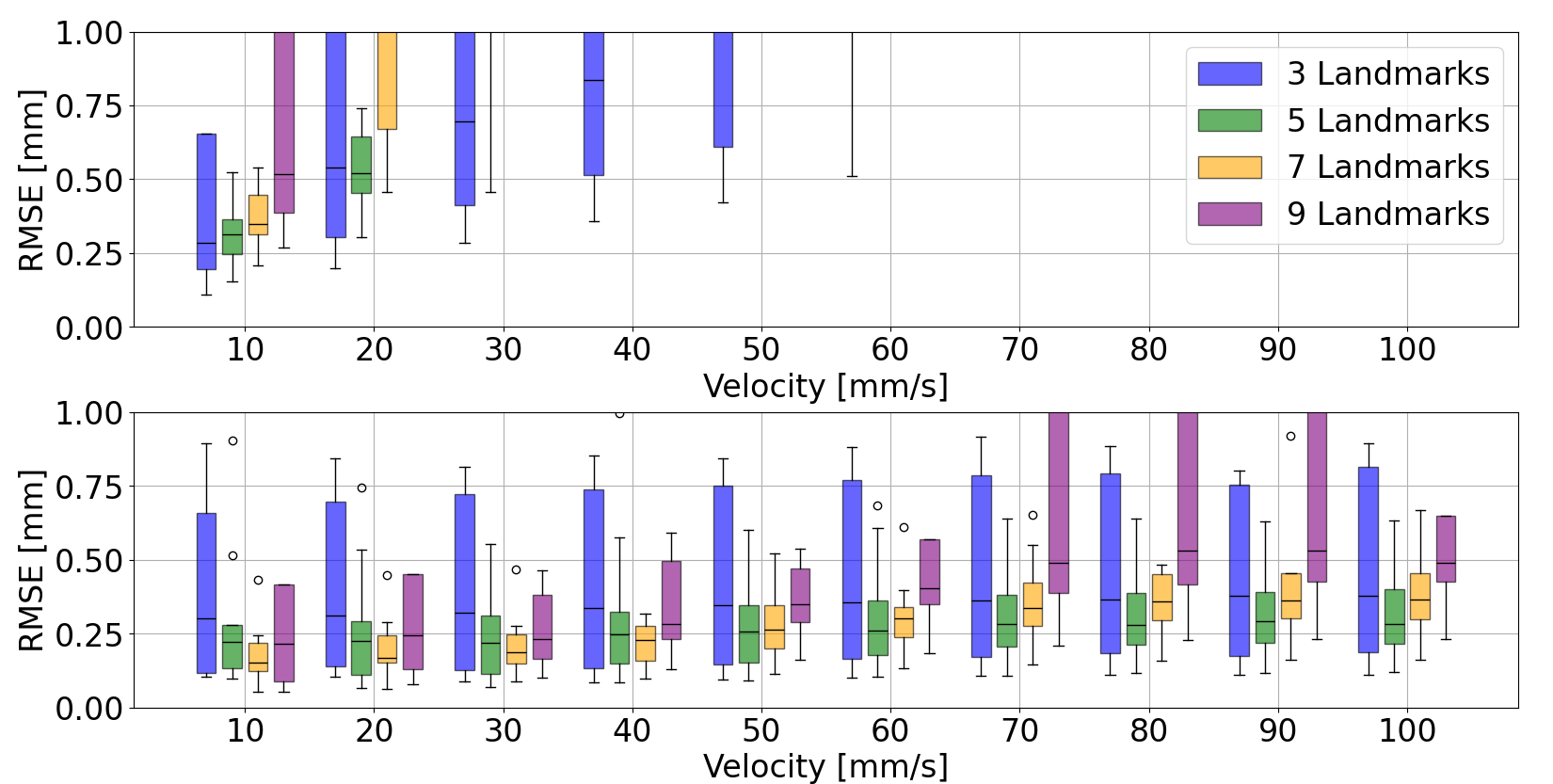}
\caption{Tracking errors for translational motion using one MOSSE instance per landmark (top) and our proposed method (bottom).}
\label{fig:old_method}
\end{figure}

Figure \ref{fig:old_method} shows the tracking performance of the baseline method (top) in comparison to the proposed method (bottom). For the baseline, tracking errors increase both with the number of simultaneously tracked landmarks and with increasing velocity. The proposed tracking method substantially increases the range of trackable velocities, with reliable tracking remaining possible at velocities of up to 100~mm/s, independent of the number of landmarks. Only for 9 landmarks, errors start to exceed 1~mm at higher velocities. In general, tracking using 5 and 7 landmarks yields lower errors compared to using 3 or 9 landmarks. The standard MOSSE achieved an inference time of approximately 1.3 ms on an Nvidia RTX A6000 GPU, while our method approximately doubles this to 2.7 ms.

\section{Discussion}

The analysis of the maximum trackable shifts demonstrates that reliable tracking with a standard MOSSE tracker is possible for lateral displacements of up to approximately 11 voxels, effectively defining the required accuracy of the proposed position prediction strategy.

As expected, the errors of the baseline approach quickly increase with increasing velocities, as the displacement between consecutive landmark observations grows. Additionally, increasing the number of simultaneously tracked landmarks decreases the maximum trackable velocity. Since landmark observations are acquired sequentially, the time between updates of a given landmark increases with the number of tracked landmarks. Consequently, the displacement between observations exceeds the allowable threshold at progressively lower target velocities.

The proposed method mitigates this limitation by propagating positional updates from a single landmark observation to all remaining landmarks via a shared global pose estimate and predicting all landmark positions at the next update step. This substantially increases the range of trackable velocities and enables successful tracking across all evaluated speeds and numbers of landmarks.

Notably, tracking with 5 and 7 landmarks consistently yields lower errors than tracking with 3 or 9 landmarks. For 3 landmarks, individual tracking errors have a stronger influence on the global pose estimate, as the global position is computed as the centroid of all tracked landmarks. Increasing the number of landmarks improves robustness against individual tracking errors, but also increases the time interval between consecutive measurements of the same landmark. As a result, motion information from previously observed landmarks becomes increasingly outdated and less reliable. This may explain the increased errors observed for 9 landmarks.

Although results are promising, they were generated for simulated motion with well-controlled dynamics. Future work will therefore focus on evaluating the proposed framework on the physical OCT tracking system. In addition, the current implementation implicitly models motion velocity in voxels per update. Since the temporal interval between updates varies depending on the spatial distance between landmarks, explicit temporal modeling may further improve robustness and prediction accuracy.

\section{Conclusion}

In this work, we present an approach for predictive motion estimation using multi-landmark OCT tracking. By employing the Kabsch algorithm to estimate global pose changes of the tracked structure, tracking robustness is improved, enabling reliable tracking even under rapid motion. Overall, the presented results are promising, but the applicability of the approach in a real-world tracking system remains to be demonstrated.

\textsf{\textbf{Author Statement}}

This research was co-funded by the MARLOC project (DFG, grant SCHL 1844-10-1) and by the European Union under Horizon Europe programme grant agreement No. 101059903; and by the European Union funds for the period 2021-2027. Informed consent: Informed consent has been obtained from all individuals included in this study. Ethical approval: not applicable.


\end{document}